%% file: main.tex
\documentclass[10pt]{article}
\usepackage[preprint]{tmlr}
\usepackage{graphicx,amsmath,amssymb,booktabs,array,placeins,longtable,flafter}
\usepackage{hyperref,url}
\graphicspath{{figures/}}

\hypersetup{hidelinks,pdfauthor={Mranmay Shetty, Gabriel Kreiman},pdftitle={Déjà View: A Shared Working-Memory Model for Visual Tasks}}
\title{Déjà View: A Shared Working-Memory Model for Visual Tasks}
\author{%
  \makebox[\textwidth][c]{%
    \begin{tabular}[t]{@{}c@{}}
      {\name Mranmay Shetty\textsuperscript{1}\quad Gabriel Kreiman\textsuperscript{2,3}} \\
      {\addr \textsuperscript{1}\,Center for Neuroscience and Cognitive Systems, Italian Institute of Technology, Rovereto, Italy} \\
      {\addr \textsuperscript{2}\,Harvard Medical School, Boston, USA} \\
      {\addr \textsuperscript{3}\,Center for Brains, Minds, and Machines, MIT, Cambridge, USA}
    \end{tabular}%
  }%
}
\begin{document}
\raggedbottom
\setcounter{topnumber}{3}
\renewcommand{\topfraction}{0.95}
\renewcommand{\textfraction}{0.05}
\renewcommand{\floatpagefraction}{0.85}
\maketitle
\begingroup\renewcommand\thefootnote{}\footnotemark\footnotetext{The authors created the task design and model, developed the scientific approach, and wrote most of the original manuscript and code. LLMs assisted with implementation and grammar correction.}\endgroup
\input{paper}
\FloatBarrier
\bibliographystyle{tmlr}
\bibliography{references}
\FloatBarrier
\appendix
\input{supplement_body}
\FloatBarrier
\end{document}

%% file: paper.tex
\begin{abstract}
A working-memory model must remember stimuli and use its memory flexibly to perform different tasks. Here we introduce a model of working memory for images that combines dynamic activity with temporary associations between image features. We train the model on five tasks, including remembering a sample while other images are presented, categorizing distractor images while remembering the sample, and determining whether an image is new or has been presented before. The same trained model performs all tasks, guided by explicit cues about which response is required. We use this flexible memory model to ask how delay, distraction, and image history affect performance. Delays have little effect on sample-memory performance for the delays we tested, while seeing more images leads to more false alarms even when the target image appeared equally recently in all cases. On the sample-memory-plus-categorization task, the probability of correct responses on all questions decreases as more questions are asked, despite robust sample-memory performance. Finally, the model maintains stable performance across 40 brief trials of each task under the tested conditions. Together, these results demonstrate how a single working-memory system can be flexibly used to perform different tasks, and how its errors depend on both previous images and task demands.
\end{abstract}

\section{Introduction}
Imagine that a computational model is shown a sequence of three images: a cat, a dog, and then the same dog. For a task to remember the first image, the correct response to the last image would be NO; in a task to detect repeated images, the correct response would be YES. Thus, a working-memory model must not only remember images but flexibly use this memory to solve different tasks. Here, we train a single model on five working-memory tasks. On some tasks, it must remember a cued sample image while other images are presented. On others, it must determine whether the current image has been seen before. On one task, it must categorize distractor images while remembering the sample for a final matching response. We use explicit task cues to instruct the model on what response is appropriate.

Persistent or dynamic activity and rapidly changing associations have been proposed as potential mechanisms of working memory~\citep{masse2019,associative,delta}. WorM has trained recurrent neural networks and Transformers on a suite of working-memory tasks and compared their performance to humans~\citep{worm}, and HebbFF has used learned synaptic plasticity to perform continual familiarity detection~\citep{hebbff}. Our model builds on these previous efforts, and our tasks unify aspects of previous tasks, such as matching a test to a sample, processing other images during a memory delay, and detecting repetitions. By using the same model on different tasks, we can ask how task demands impact model behavior.

Here we first introduce our tasks and flexible working-memory model. Next, we ask how performance changes with increasing delays and image histories, and across 40 short trials. Finally, we examine how activity and associations contribute to memory by comparing our full model to models trained to use only one type of memory, as well as to two other architectures.

\subsection{Contributions}
\begin{itemize}
\item We introduce a flexible working-memory model that can be cued to perform five different memory tasks.
\item We show how dynamic activity and rapidly changing associations contribute to performance, with benefits that depend on the task.
\item We characterize how delays, distractors, and previously seen images impact performance on different memory tasks.
\end{itemize}

\section{Task design}\label{sec:tasks}
\begin{figure}[htbp]
\centering\includegraphics[width=\textwidth]{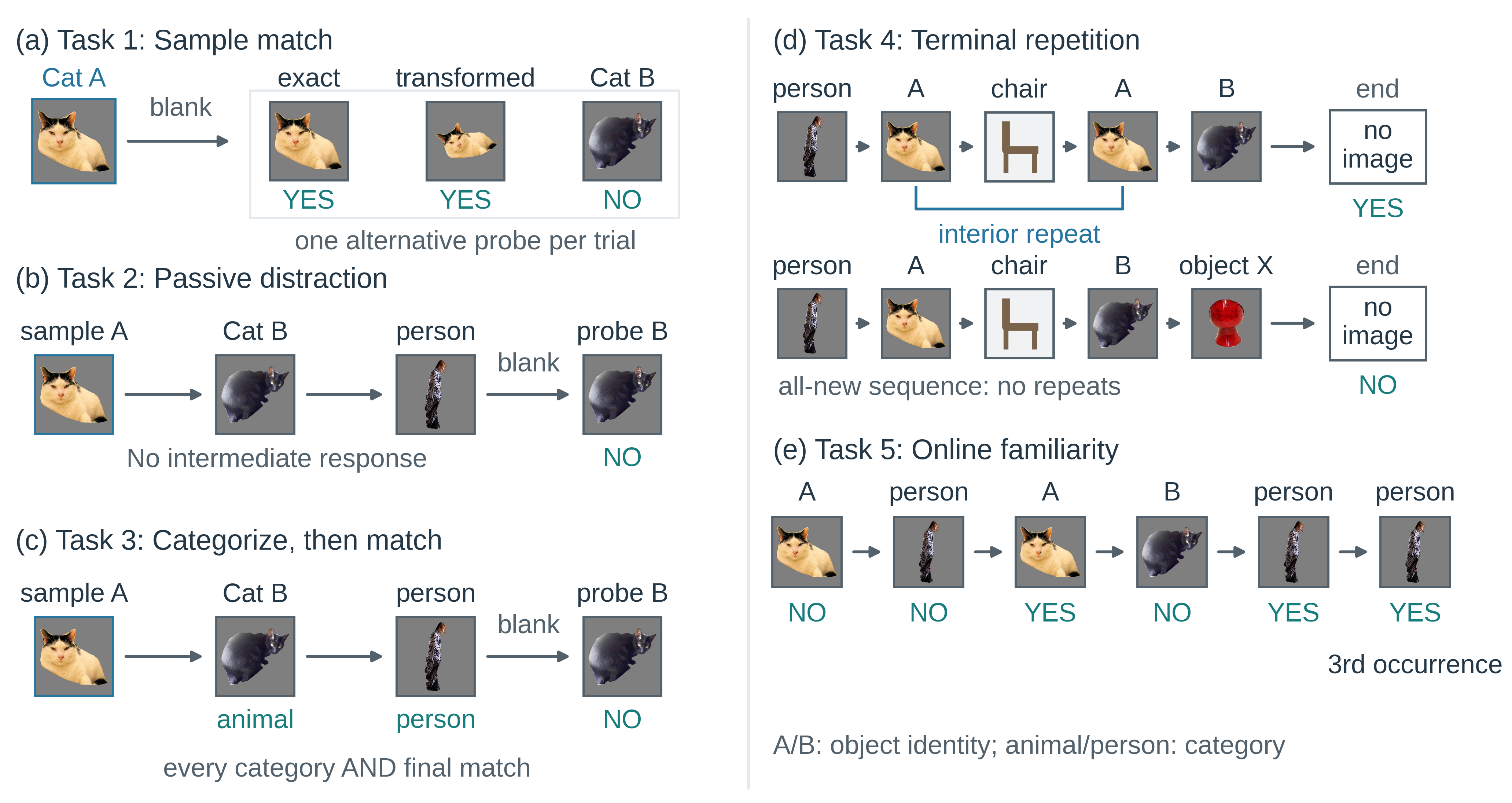}
\caption{Tasks performed by our working-memory model. On Tasks 1--3, the model must report whether the test image matches the cued sample image. On Task 3, it must also categorize the distractor images presented during the delay. On Tasks 4--5, it must report whether any image repeats (Task 4) or whether each image has been presented before (Task 5). Responses shown are ground truth, not model responses or inputs. The transformed example is an altered version of the same photograph. }\label{fig:tasks}
\end{figure}

Our five working-memory tasks use foreground images of objects on neutral backgrounds, with blank intervals between images (Figure~\ref{fig:tasks}). On Tasks 1--3, the model must remember a cued sample. On Tasks 4--5, it must determine whether images have appeared earlier in the current trial.

\paragraph{Task 1: Remember the sample.} The model sees a sample image, a delay with no image, and then a test image. It must report whether the test image matches the sample.

\paragraph{Task 2: Remember through distraction.} Like Task 1, but other images are presented during the delay and the model must not respond to them. Importantly, if the final test image is one of these distractors, the correct response is still NO because this image is not the sample.

\paragraph{Task 3: Categorize while remembering.} Like Task 2, but the model must also categorize each image presented during the delay. We score these categorization responses separately from sample-memory performance. Complete-trial success requires every category response and the final sample-memory response to be correct.

\paragraph{Task 4: Report repetition at the end.} The model sees a sequence of images and does not make responses during the sequence. At the end, it must report whether any image repeated. There is no cued sample image, and the final question is not associated with an image.

\paragraph{Task 5: Recognize each repetition.} The model must make a response after each image, reporting whether it has been presented before in the trial. Each image's first appearance requires NO, and later appearances require YES. The model makes this response before the image is added to its memory.

Importantly, on all tasks images from the same category but depicting different objects, such as two different chairs, are not considered matches. The model receives cues specifying the task, the requested response, timing and image count, trial context, and the role of each image. It does not receive object identities, category labels, future sequence length or correct answers. Each trial starts a new relevant history, although internal memory can carry over from the previous trial. Thus, an image seen on a previous trial can be familiar without counting as a repetition in the current trial. The delay and image-history experiments use repetitions of identical images. The forty-trial sessions can include different renderings of the same source object. Neither study tests recognition across different photographs of the same object.

\section{The shared memory model}\label{sec:model}
\begin{figure}[htbp]
\centering\includegraphics[width=\textwidth]{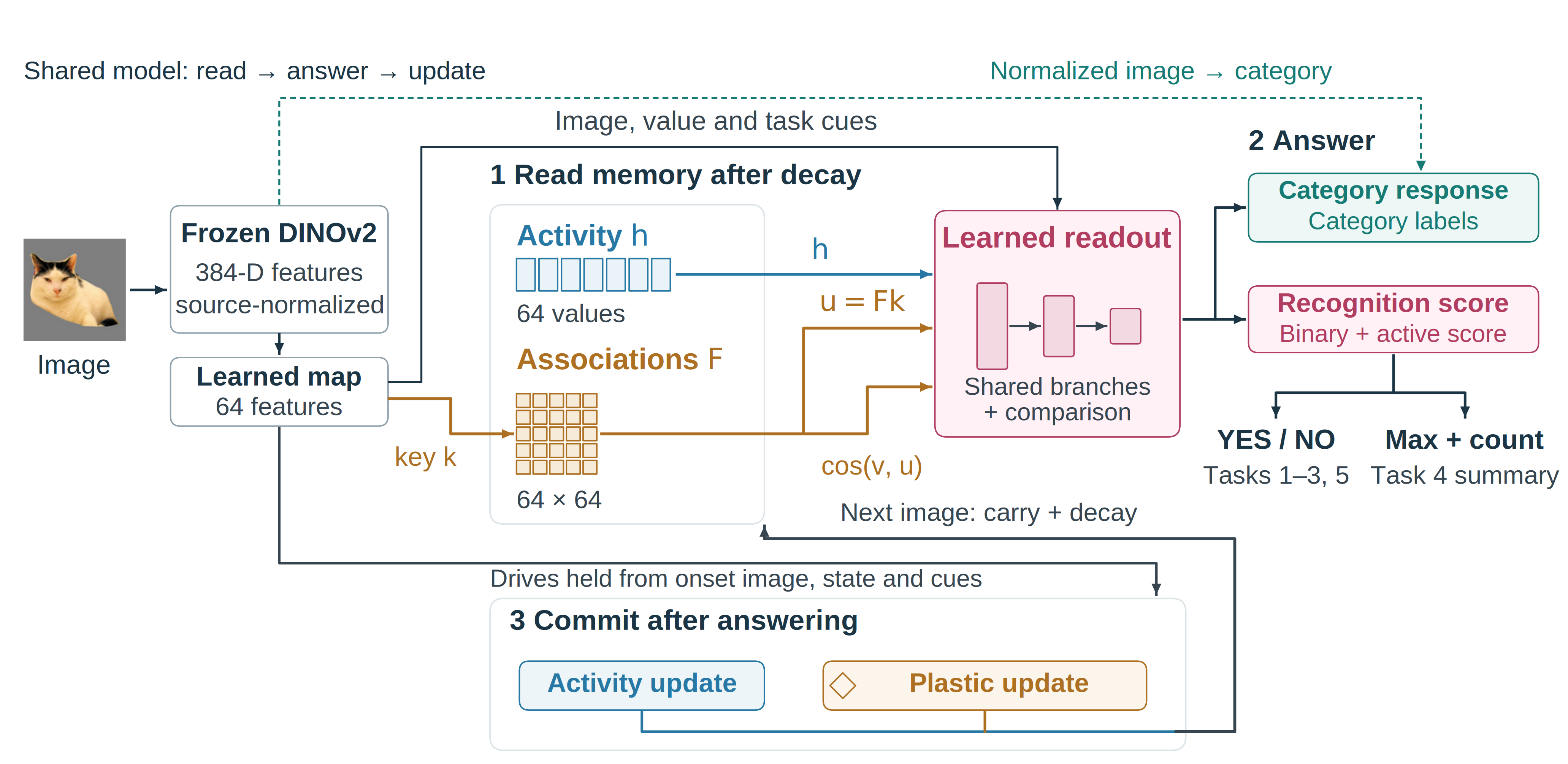}
\caption{Schematic of our working-memory model. Activity ($h$) and associations ($F$) change over time. Context and role cues modify the keys that address the associations. Responses combine activity, retrieved content, image features and cues; category responses also receive normalized image features directly. The model answers from aged memory before writing the current image. Task 4 uses the largest local recognition score and image count for its final answer.}\label{fig:model}
\end{figure}

Our model has two types of rapidly changing, temporary memories: an activity state, $h\in\mathbb{R}^{64}$, and associations, $F\in\mathbb{R}^{64\times64}$ (Figure~\ref{fig:model}). Activity changes based on the current image and information read out of the associations, which link keys (inputs used to access memories) with values (information representing image content). Context and role cues modify the key used to access memories, so that the same image can be remembered differently under different instructions. Throughout a trial, $h$ and $F$ change, but the model's learned parameters remain fixed.

\subsection{Encoding images}
We used a pretrained, frozen DINOv2 encoder~\citep{dinov2} to transform images into 384-dimensional feature vectors. These features were normalized to unit length and then standardized using statistics from the training data. A learned linear layer followed by a tanh activation transformed them into 64-dimensional image representations.

\subsection{Responding}
The model's key for the current image is used to read a value out of the associations. The model's recognition score uses this value, the activity state, features from the current image, and public cues. For this response, both memory states are aged through the image's exposure before they are read. We also provide the model with ``retrieval cosine'': the cosine similarity between the retrieved value and the current image's value, which compares their directions while ignoring their magnitudes. Tasks 1--3 and 5 use this shared score with their respective cues and calibrated response rules. On Task 3, category responses use the shared reader inputs plus a direct pathway from standardized image features. On Task 4, the final response combines the maximum unthresholded local recognition score over the trial with a learned adjustment for image count. This response follows a question cue with no image. The same network serves all five tasks; we did not train five separate models.

\subsection{Updating memories}
Both memory states decay as time passes. When an image appears, the current image, activity, retrieved content and cues determine an activity drive and update rates, which stay fixed during its exposure. Activity moves toward this drive while decaying. Associations are updated so that the value retrieved by the image's key moves toward the current image's value~\citep{delta}. The model answers from memory aged through the exposure before committing these updates. Its predicted answer does not determine what is written. Appendix~\ref{supp:model} summarizes the implementation.

\subsection{Training}
We used COCO images~\citep{coco}, splitting the photographs into sets used to train the model's weights (1,920), choose between checkpoints (480), calibrate its responses (480), and assess performance during development (960). We trained three models with different random initializations for 6,000 gradient updates. Half of the training objective covers Tasks 1--5, including category responses on Task 3 and local repetition labels as well as the final answer on Task 4. The other half teaches context membership, designated-item recognition and role recognition.

We chose checkpoints based on loss on photographs used only for model selection, and calibrated probability estimates and YES/NO rules using a separate set of photographs. The selected weights and calibrated response rules remain fixed in all experiments below. Appendix~\ref{supp:training} gives the training ranges, objective weights and selection procedure.

\section{Experiments}\label{sec:behavior}
We used the development images and all three models described above for these controlled experiments. On Task 1, we kept the sample and query images fixed while varying the blank delay. On Tasks 2 and 3, we kept the sample, query and time to the answer fixed while varying the number of distractors. On Tasks 4 and 5, we kept the target and final query images fixed while varying the number of images. Balanced accuracy gives equal weight to the rates of correct matching and nonmatching answers, or repeat and new-image answers. In Task 3, complete-trial success requires every category answer and the final memory answer to be correct.

\begin{figure}[htbp]\centering\includegraphics[width=\textwidth]{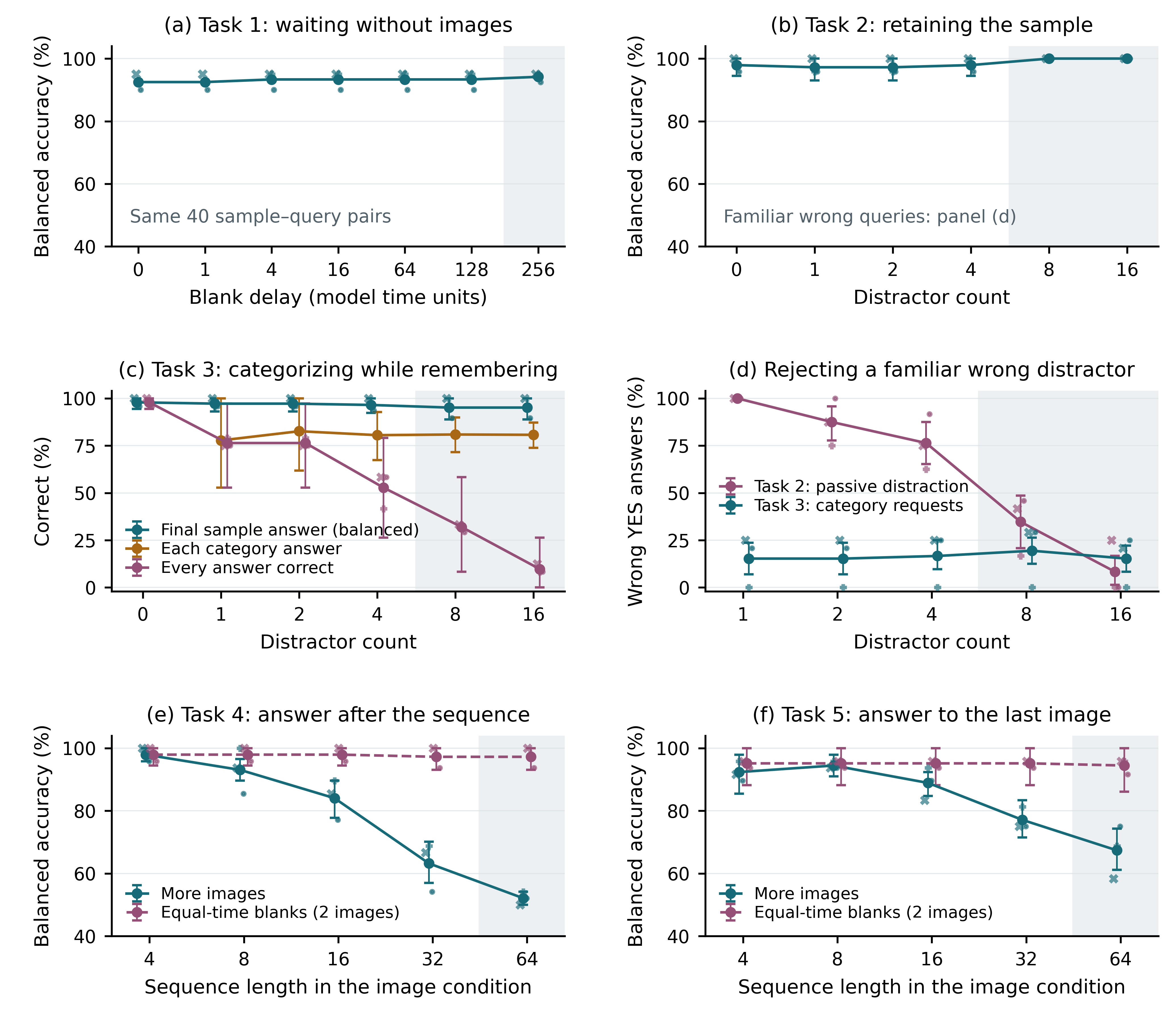}
\caption{The same pictures under increasing memory demands. (a) Task 1: forty fixed sample--query pairs across blank delays. (b,c) Sample matching with novel nonmatching queries, with sample-response age fixed at 65 time units. (c) Category answers and complete-trial success are scored separately. (d) Familiar distractors used as wrong queries. (e,f) Longer image sequences compared with equal-duration blanks containing only two images; Task 5 shows the last response. Means average three models. Bars are exploratory 95\% intervals over twelve photograph groups; (a) is descriptive. Small markers show model means except for category accuracy. Gray regions exceed training support. Lines connect the tested conditions, spaced equally for display.}\label{fig:behavior}\end{figure}

\subsection{Effects of delay and image history}

For Task 1 (Figure~\ref{fig:behavior}a), mean performance (balanced accuracy across 40 sample--query pairs) without a delay was 92.5\% and with a delay of 256 model time units was 94.2\%.

Memory states decay during a blank delay interval, but balanced accuracy remained high over the tested delays. While decay leads to a uniformly scaled associative memory matrix, for a given query this has no effect on the direction of memory retrieval, provided retrieval remains nonzero. Thus similarity information can be maintained even as the overall strength of memory decreases, although this alone does not explain the model's responses.

For Tasks 2 and 3 (Figure~\ref{fig:behavior}b,c,d), sample-matching performance was high (balanced accuracy above 97\% and 95\% in Tasks 2 and 3, respectively) when the incorrect query was a novel image. When the incorrect query was a familiar distractor image, performance varied: in Task 2 the model responded YES to all incorrect queries when there was one distractor, and to 8.3\% when there were 16; in Task 3 the model responded YES to 15.3\% when there was either one or 16 distractors. Adding more distractors did not uniformly degrade performance, even though the identity and elapsed time since presentation of the incorrect query image were held constant.

In Task 3, accuracy for individual category responses remained relatively high (about 80\%), but the percentage of trials with correct responses to all category questions and the sample-matching question decreased as the number of distractors increased (76.4\%, 31.9\%, and 9.7\% for 1, 8, and 16 distractors, respectively; Figure~\ref{fig:behavior}c). Sample-matching performance in isolation remained high. Performance on the full trial was directly measured, not estimated by multiplying accuracies.

For Tasks 4 and 5 (Figure~\ref{fig:behavior}e,f), performance decreased as the number of images increased from 4 to 64 (balanced accuracy on the terminal question decreased from 97.9\% to 52.1\% in Task 4, and balanced accuracy on the last query decreased from 92.4\% to 67.4\% in Task 5). Interestingly, in Task 5 performance was slightly higher after seeing 8 images compared to after 4. False-alarm rate increased from 4.2\% to 91.7\% in Task 4 and from 15.3\% to 51.4\% in Task 5 as list length increased from 4 to 64 images.

When we presented blank delays instead of images between the target and query images (matching the total amount of time between these two images), performance remained high (balanced accuracy of 97.2\% and 94.4\% for the longest duration in Tasks 4 and 5, respectively). This indicates that the passage of time alone does not explain the decrease in performance, although this change also affected the number of memory updates and images seen, and in Task 4 also affected the information used for answering the question. Thus we cannot isolate which of these changes caused the impaired performance with images. In these experiments only the last image could be repeated, so in Task 5 we report performance on the last query, not averaged across the sequence.

\subsection{Earlier history with equally recent targets}

\begin{figure}[htbp]\centering\includegraphics[width=\textwidth]{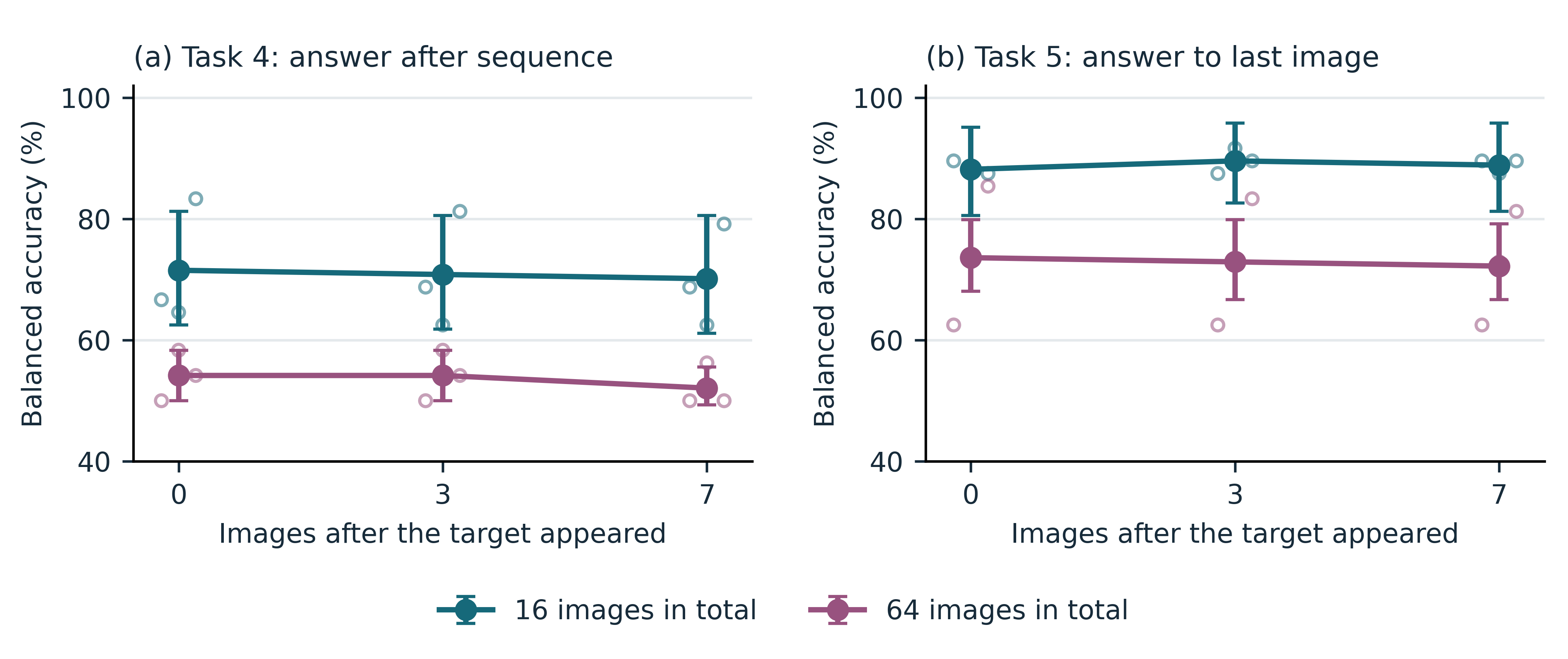}
\caption{Longer histories with equally recent targets. The target, query and images between them were identical across lengths. The longer sequence contained 48 additional unique images before the target; neither condition added a blank delay. Means average twelve photograph groups and three models. Open points show individual model means, and bars show descriptive 95\% intervals over photograph groups. Length 64 exceeds training support. Appendix Table~\ref{supp:factorial_contrasts} reports the two length comparisons with intervals adjusted across all nine planned comparisons.}\label{fig:recentness}
\end{figure}

In Tasks 4 and 5 increasing the number of images between the target and query images also increased the amount of time since the target image was seen. To tease apart these effects we compared performance on sequences of 16 and 64 images which were identical from the target image onward (with either zero, three, or seven images presented between the target and query images; the 64-image sequence contained 48 additional novel images before the target image). Averaged across these conditions, balanced accuracy decreased from 70.8\% to 53.5\% in Task 4 and from 88.9\% to 72.9\% in Task 5 as sequence length increased from 16 to 64 images (Figure~\ref{fig:recentness}), and these effects remained significant after correcting for the nine planned comparisons. Hit rate for repeated images remained high (98--100\%) while false-alarm rate increased from 56.5\% to 91.7\% in Task 4 and from 22.2\% to 52.3\% in Task 5, indicating that novel images seen earlier in the sequence can lead to false familiarity even when controlling for the recency of the target image. This controls target age, but changes the number of images and memory updates, the memory state before the target and, in Task 4, the information accumulated for the final answer. We therefore cannot tease apart these potential causes of impaired performance.

We also varied blank delay alongside sample exposure (Task 1), distractor count (Tasks 2 and 3), or sequence length (Tasks 4 and 5). The added delay occurred after sample presentation in Task 1, after the distractor schedule in Tasks 2 and 3, and before the final query image in Tasks 4 and 5, not before Task 4's image-free answer. Added blank delay is distinct from the total time since the sample or target image, which can also include intervening images. None of the seven planned comparisons assessing an interaction with delay established an effect after correction across all nine planned comparisons (see Appendix~\ref{supp:factorial} for all planned comparisons and a full grid of results).

\subsection{Forty short trials}

In order to assess performance over the course of a session, we presented forty trials in each session, using cyclic rotations such that each of forty fixed trials appeared in each of forty trial positions. With memory carried between trials, performance changed little from the first ten to last ten trials (range of changes across nine fixed conditions spanning all five tasks: $-0.08$ to $+0.50$ percentage points; see Appendix Table~\ref{supp:fixed40} for full results). This is a descriptive range of performance changes, not an average change or a significance test. Thus, even when performance on an individual longer trial may be impaired, overall performance across a session made up of shorter trials can remain stable. This does not mean that the model can successfully retain a memory of one sample image for all forty trials; each trial is a distinct bounded task, and these experiments assess repeated use rather than retention of the same sample across the session.

\section{What do the two memories contribute?}\label{sec:comparisons}

We next sought to better understand the contributions of each type of memory by training models with only one type of learned memory: either only the activity state or only the associative memory matrix. We also trained a causal Transformer and a Gated DeltaNet model. All models were trained to perform the same set of tasks with the same number of training updates, but were not otherwise matched for memory size, compute, or convergence. Table~\ref{tab:samebattery} shows results across a wider range of delays, sequence lengths, and exposures than in previous sections, including some conditions outside of training support.

\begin{table}[htbp]\centering
\caption{Performance across the broad task battery (\%). Tasks 1, 2, 4 and 5 report balanced accuracy across responses; Task 3 reports the percentage of trials with every response correct. Values average three trained models, using each model's source-calibrated response rules and resetting memory between trials. Activity-only and associations-only models were trained separately. Equal training-update budgets do not match memory capacity, computation or convergence.}\label{tab:samebattery}
\input{tables/SAME_PROTOCOL_MAIN.tex}\end{table}

Different types of memory were most beneficial for different tasks. The model with only activity-based memory had the highest mean performance on Tasks 1 and 2, while the model with only association-based memory had the highest mean performance on Tasks 3 and 4, and exceeded the full model's mean on four out of five tasks. For Task 5, performance of the full model was 12.48 percentage points higher than the model with only activity-based memory (exploratory 95\% interval: [11.02, 13.78]), while the full model exceeded the model with only association-based memory by 0.86 percentage points (exploratory 95\% interval: [0.41, 1.35]). Mean performance on Task 3 (fraction of trials on which all responses were correct) was higher for the full model than for both the Transformer and Gated DeltaNet models, but the Transformer had the highest mean performance on Task 5. These comparisons concern models trained separately without a component, rather than components removed after training.

\section{Discussion}

We introduced a model that can memorize sample images, categorize other images, and make repetition judgments according to different task demands. Performance was impaired by additional images presented prior to the target (Tasks 4 and 5), and complete-trial success fell as the number of images to categorize increased (Task 3), while sample matching changed little over the tested delays. Other aspects of performance proved more robust, including sample memory in Task 3 and performance over a session made up of many short trials. We also demonstrated that temporary associations between image features can play an especially important role in supporting online repetition detection, and that different forms of memory make different contributions across these tasks.

Our model could be useful for generating testable predictions for experiments that record from the brain. Prior work has investigated the potential contributions of neural activity and short-term synaptic plasticity to working memory~\citep{masse2019}. In particular, \citet{kozachkov2022} compared models with and without synaptic plasticity to neural activity recorded from prefrontal cortex while a non-human primate performed an object-memory task that included distractors. Our model could be used in a similar manner: neural activity recorded prior to, during, and after presentation of a distractor could be compared to activity from our full model and the model trained without associative memory, using matched stimuli, task demands, and timing. With model performance also taken into account, such an analysis could test on held-out trials whether associations help explain neural activity beyond differences in task accuracy. A separate behavioral experiment could test the prediction that more images presented prior to the target increase false familiarity even when target age is held fixed. While neural recordings may distinguish models with and without associations, they would not provide a direct readout of the associations stored in $F$.

There are many potential extensions to our model. For instance, our model was provided with explicit information about task, trial boundaries, and time; future work could investigate how performance changes when such information is less explicit. Our model computed a decision variable for Task 4 using the largest local recognition score and an adjustment for image count; future work could investigate alternative strategies. Although we probed model performance on some sequence lengths and numbers of distractors that exceeded those seen during training, future work could investigate learning and performance on a wider range of task demands.

%% file: tables/SAME_PROTOCOL_MAIN.tex
\begin{tabular}{@{}lrrrrr@{}}\toprule
System & T1 & T2 & T3 joint & T4 & T5 \\ \midrule
Shared h/F & 90.71 & 87.61 & 44.90 & 67.03 & 76.75 \\
Activity only & 91.84 & 91.61 & 43.15 & 63.85 & 64.28 \\
Associations only & 91.58 & 88.27 & 45.61 & 69.32 & 75.89 \\
Transformer & 90.62 & 88.38 & 40.57 & 67.19 & 83.29 \\
Gated DeltaNet & 90.62 & 89.53 & 43.26 & 60.42 & 69.22 \\
\bottomrule\end{tabular}

%% file: supplement_body.tex
\section{Data and evaluation}\label{supp:data}

Unless otherwise specified, stimuli in behavioral experiments consisted of masked foreground objects presented on a neutral background on a $224\times224$-pixel canvas. The image encoder and normalization statistics were fixed prior to model evaluation. We used disjoint sets of COCO photographs for training model weights (1,920), selecting model checkpoints (480), calibrating models (480), and development evaluation (960). We counted unique photographs rather than crops, renderings, or copies used as queries. For the experiment investigating readout of a fixed memory (Appendix~\ref{supp:ordering}), we used 32 groups of images from Open Images~\citep{openimages} that we had previously used in other experiments, for a total of 1,024 photographs and 1,056 selected object instances. Each group included 31 support photographs and two anchor objects from the same category depicted in one additional photograph. These photographs were distinct from those used to fit readouts. Both the behavioral and same-memory assessments are exploratory because the photographs had already been examined. We did not establish whether the sets contained near duplicates or additional photographs of the same physical object, or whether any images overlapped with data used to pretrain the image encoder.

Response age was defined as the time, in model units rather than milliseconds, from the offset of the earlier image through the end of the query exposure. Image lag was defined as the difference in image positions between the query and its most recent matching presentation; consecutive presentations have lag one. We report hit rates (the proportion of repeat or sample-match queries, depending on the task, that were endorsed) and false-alarm rates (the proportion of new or nonmatching queries, depending on the task, that were endorsed). Balanced accuracy is the average of hit rate and correct-rejection rate. For Task 3, we also report the proportion of trials on which the model made correct categorization judgments for all distractor images and a correct sample-memory judgment (complete-trial success); when no distractors were presented, this metric reduces to final sample-memory accuracy and categorization accuracy is undefined. We only included trials for which the model made all required responses. The reported confidence intervals describe the fixed trained models and do not include uncertainty from retraining or checkpoint selection.

For the battery of tasks in Table~\ref{supp:battery}, we cleared memory between trials and pooled binary responses within each task, rather than scoring only the final query or averaging balanced accuracy across streams. Task 3's complete-trial endpoint was scored separately. Accuracy on individual categorization responses, accuracy on all categorization responses, and complete-trial success are different metrics.

\begin{table}[htbp]\centering\small
\caption{Response counts per trained model. Tasks 1--4 have one final binary answer per trial; Task 3 adds 2,336 category answers. Task 5 has 1,792 answers in the length panel, 4,096 in 512-image streams and 3,072 in the exposure panel. Exposures are 0, 0.25, 1 or 4 units, with fixed-interval or matched-duration schedules.}\label{supp:battery}\label{supp:support}
\input{tables/BATTERY_PROTOCOL.tex}\end{table}

\section{Model specification}\label{supp:model}\label{supp:exact}

The response model has 96- and 64-unit hidden layers and a comparison branch that learns its weights. Its inputs are current-image features, activity, retrieved content, their interactions and similarities, activity, retrieval and association norms, and public cues. The response model's image input allows it direct access to the current image; exposure duration scales the effect of this input. Category responses also use a direct image pathway with a 96-unit hidden layer and 12 outputs.

During blanks, activity and associations decay exponentially. When an image appears, the model computes a bounded target activity and activity and association update rates using the image features, previous activity, retrieved content and public cues; these values do not change for the duration of the exposure. During exposures, activity relaxes toward its target and decays; associations decay and adjust retrieved content toward the current image's value vector. When forming a response, the model first ages its memory through the exposure duration, then reads activity and retrieved content before writing the current image. This retrieval differs from the retrieval at image onset. The model does not use its predicted response when updating memory. These rules define a transition that integrates decay and writing, given fixed update targets and rates.

At trial boundaries, the model advances its clock and ages memory, optionally resets or halves activity and associations, clears its decision summaries and advances the context cue. Half retention multiplies both memory components by one-half in addition to normal decay. For the first image in each trial, activity and associations are multiplied by learned task-dependent factors between zero and one. For the final response in Task 4, which does not depend on a current image, the model uses a learned scaling of the maximum local recognition score (without thresholding) and a learned adjustment based on how many images it has seen. All neural reference models use this final response procedure.

\section{Training and reference models}\label{supp:training}

We trained Tasks 1--5 with loss weight 0.1 each. Context membership, designated-item recognition and item-role recognition each had weight one-sixth. Binary responses used binary cross-entropy and category responses used 12-class cross-entropy. Within Task 3, we added final memory loss to mean category loss. Within Task 4, we added final response loss to one-quarter of mean local repetition loss, including nonrepeats. All other tasks used mean binary loss. We averaged over trials within each task, not questions, so adding questions does not down-weight Task 3's final memory loss.

For context exercises, we separately averaged binary and category losses, then averaged over episodes. For designated-item exercises, we weighted the positive example's loss by one-half and each of the three negative examples' losses by one-sixth. For item-role exercises, we equally weighted positive and negative losses, including wrong-role, previously unseen, neutral and prior-trial-only negatives, then averaged over role requests and role assignments. Context exercise memory loads were 1--4 items. Designated-item exercise memory loads were 2, 4 or 5 items and used 0--2 previous trials. Item-role exercise memory loads were 2 and 4 items and used both role assignments. We did not separately isolate the effects of these additional objectives.

We trained from three random initializations for 6,000 AdamW optimizer steps with learning rate 0.001, weight decay 0.0001 and gradient-norm clipping at 5. Each step used ten four-trial histories (eight trials per task), four context episodes, two designated-item templates and four item-role templates, computing gradients over whole histories without truncation. Batches for different losses were independently initialized. Eight-time-unit boundaries used reset, half retention or carry. Blank durations (0, 16, 64 and 128), numbers of distractors (0, 1, 2 and 4), Task 4/5 list lengths (4, 8, 16 and 32) and image exposure durations (0.25, 1 and 4) were sampled during training. Training configurations specify short-query augmentation; four-trial histories can exceed the single-trial training length limit. During training, we evaluated four checkpoint-selection schedules at initialization and every 500 updates. For each initialization, we selected the checkpoint with the lowest average loss without response calibration, breaking ties toward earlier updates.

We calibrated responses using held-out calibration photographs: five affine transformations with positive slopes, one each for Tasks 1--3 (sample), online repetition, final response, designated-item and item-role responses. Calibrated logits of zero or above predict YES. Categories have a positive temperature calibration that does not change which category has maximum probability. These transformations are fixed during evaluation. We used L-BFGS-B to minimize source-weighted binary cross-entropy, with source weights summing to one for each response policy. We used 1,873 feature rows from 479 of the 480 calibration photographs.

For ablations, we trained three models per component (activity and associations) from random initialization using the same procedure as above. Activity-only models zero associations before responses, updates and trial boundaries and disable association writes; associations-only models zero activity and its target. The remaining parameters were trained. Both retain image inputs, public cues, category responses and final summaries; they differ in capacity.

\paragraph{Reference neural network models.}
The Transformer has two layers of width 128 with four attention heads and feed-forward layer width 256. Gated DeltaNet~\citep{gateddeltanet} has two layers of width 128 with three attention heads, key and value widths 32 and 64, and width-4 convolutions; we use the model class from FLA with PyTorch reference operators. We trained three models per architecture from random initialization using the same losses, number of training steps and checkpoint-selection procedure. Unlike our model, reference models' memories do not decay during blanks, although their clocks still advance. Unlike our model with fixed-size memory, the Transformer's key/value memory grows with sequence length. Gated DeltaNet has fixed-size convolutional and recurrent memories. We did not match model size, computation or convergence.

\paragraph{Uncertainty intervals.}
We computed exploratory 95\% uncertainty intervals for component comparisons by resampling blocks of generated trials with replacement 2,000 times, stratified by protocol and using the same draws for all models. Blocks can contain repeated photographs. These intervals describe uncertainty over the fixed evaluation protocol, conditional on the trained models; they do not include retraining uncertainty.

\section{Controlled task results}\label{supp:tasks}\label{supp:controlled}
\paragraph{Controlled experiments.}
For Task 1, we used forty fixed sample--query pairs, twenty matches and twenty nonmatches. For each of seven blank durations (0, 1, 4, 16, 64, 128 or 256), we copied the memory state after the sample and evaluated each copy independently. Images were exposed for one time unit.

For Tasks 2--5, we used twelve photograph-disjoint groups of 64 COCO development photographs. Fixed hashes selected one object instance and rendering per photograph, two images in the same fine-grained category (anchors), and 62 other images (fillers) within each group, without using model responses. Selected photograph and rendering hashes were unique, but the overlap limitations in Appendix~\ref{supp:data} still apply.

For Tasks 2/3, after the sample image we presented sixteen possible four-time-unit slots. Filled slots contained three time units of blank followed by one time unit of a distractor image; unfilled slots were entirely blank. To present 0, 1, 2, 4, 8 and 16 distractors, we filled slots in the following order: 8, 4, 12, 2, 6, 10, 14, 1, 3, 5, 7, 9, 11, 13, 15, 16. Images were presented in chronological slot order, keeping the sample-response memory age at 65 time units. After each distractor sequence and its mirror, we queried both anchors independently. A separate familiar nonmatching query tested the distractor in slot 8 at response age 33. This distractor appears at every nonzero distractor count; the familiar query is undefined at zero. In these experiments, models responded YES to all positive sample queries, so errors on novel queries were false alarms (YES responses). We counted category responses once per history, not once per query copy.

For Tasks 4/5, after an anchor image we presented a prefix of fillers, with longer sequences nesting shorter ones, then queried either anchor using separate copies of memory. List lengths were 4, 8, 16, 32 and 64 images, and only the final image could repeat. For Task 4, we computed that image's local recognition score before its memory update and committed the image before the image-free final response. For Task 5, the model forms responses at every image before writing it. For equal-duration blank controls, we replaced all images except the first (target) and final (query) images with blanks for the same total amount of time. All trials began with empty memory and an eight-time-unit trial boundary, retaining first-image attenuation. Images were exposed for one time unit. This control changes both image count and the number of memory updates.

\begin{table}[htbp]\centering\small
\caption{Class-specific repetition results (\%). Task 4 shows its image-free final response; Task 5 shows only the response to the final query image. BA averages hit and correct-rejection rates. Blank conditions match elapsed time but contain only the first and last images. All results are averaged over three models and twelve photograph groups.}\label{supp:controlled_endpoints}
\input{tables/CONTROLLED_ENDPOINTS.tex}\end{table}

Exploratory 95\% confidence intervals use 10,000 bootstrap resamples of the twelve photograph groups, retaining all conditions and all three fixed models together within each sampled group. At 64 images, image-minus-blank differences in balanced accuracy are $-45.1$ percentage points [$-49.3,-40.3$] for Task 4 and $-27.1$ [$-32.6,-21.5$] for Task 5. These intervals do not reflect uncertainty in training or model selection. Zero-width intervals when models achieve perfect scores do not establish perfect performance on all possible images.

\paragraph{Crossed delay, distraction and sequence length.}\label{supp:factorial}
We used the same twelve photograph groups and three models to evaluate 164 experimental conditions without further training or image selection. In Task 1, we crossed sample exposures of 0.25, 1 and 4 time units with blank durations of 0, 16, 64 and 256; query exposure was always one time unit. We used twelve anchor pairs, one per group, with mirrored histories and independent queries. In Tasks 2/3, we crossed the same four blank durations with six distractor counts. The added blank duration $D$ followed the complete distraction sequence, giving sample-response ages of $65+D$ and familiar-query ages of $33+D$. Category responses occur before this added delay and are identical across delay copies. In Tasks 4/5, we crossed five sequence lengths with the same four blank durations for image and blank conditions. The added delay occurred before the final image, not before Task 4's image-free response.

\paragraph{Recent targets.}
We also evaluated a smaller experiment varying added blank duration (0 and 64 time units), number of images between target and query (0, 3 and 7), and total sequence length (16 and 64 images); only the final image could repeat. Each stream used the last $N-2$ fillers in the predefined order, with the target inserted so that exactly $I$ images lay between it and the query. Target-response age was $I+1+D$ time units. The images after the target were identical and in the same order across sequence lengths. We did not change first-image attenuation, response thresholds or Task 4's controller. The public image-count cue reflects the number of images already seen, not the future sequence length. List length 64 and Tasks 2/3 with 8 or 16 distractors exceed the ranges used during training.

\begin{figure}[htbp]\centering\includegraphics[width=\textwidth]{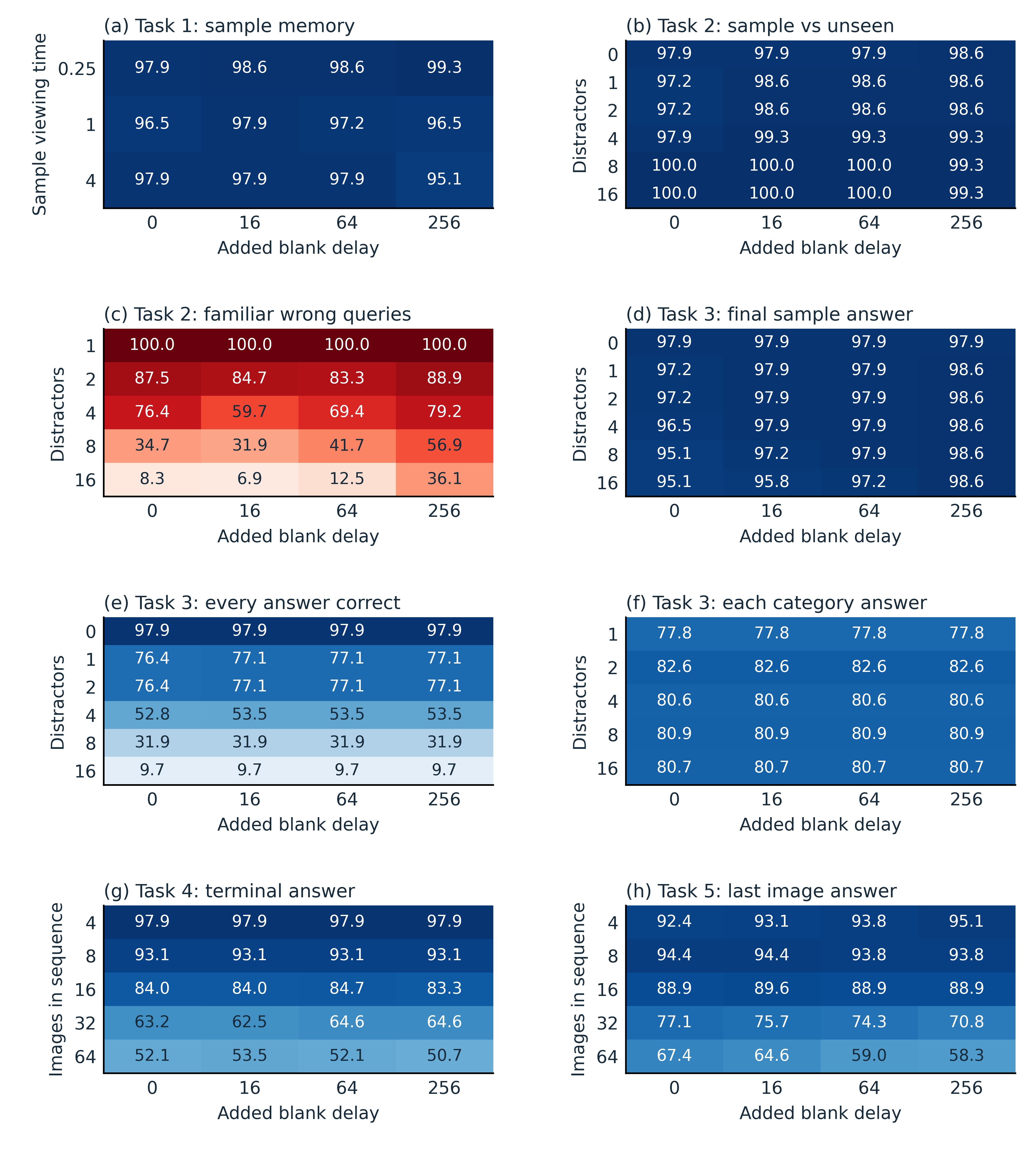}
\caption{Crossed image-history and delay tests. Values are percentages, averaged over twelve groups and three fixed models. Blue shows correct responses; red shows familiar-wrong false positives. Final memory and repetition use balanced accuracy. Category answers occur before the added delay, and are undefined at zero distractors. The nine planned comparisons are reported in Table~\ref{supp:factorial_contrasts}.}\label{supp:factorial_grid}\end{figure}

Figure~\ref{supp:factorial_grid} shows the crossed delay results; the Task 4/5 panels show conditions with intervening images. We prespecified nine comparisons for this factorial experiment (Table~\ref{supp:factorial_contrasts}). The first seven test whether the effect of sample exposure, distractor count or list length changes between zero and a longer blank delay. Task 1 compares exposures of 4 versus 0.25 at delays of 256 versus zero. Tasks 2/3 compare 4 versus 0 distractors for sample balanced accuracy; Task 2 compares 4 versus 1 for familiar-query false alarms, and Task 3 compares 4 versus 1 for complete-trial success. These comparisons use delays of 64 versus zero. Tasks 4/5 compare lengths of 32 versus 8 at delays of 64 versus zero. The final two comparisons, also in Tasks 4/5, test lengths of 64 versus 16 without added delay, equally weighting the three intervening-image counts.

We computed 50,000 bootstrap resamples of the twelve photograph groups, keeping all conditions and three fixed models together within each sampled group. Bonferroni correction for the nine comparisons gives approximately 99.444\% confidence intervals. All seven intervals for interactions with delay include zero; both intervals for sequence length at fixed target recentness exclude zero. Although the comparisons were planned before this evaluation, the photographs had already been examined, so the results remain exploratory.

Table~\ref{supp:factorial_contrasts} reports the comparisons planned before this factorial evaluation. All contrasts are within tasks. Viewing-time and count comparisons examine the interaction of delay with viewing time (Task 1) or distractor count (Tasks 2--3). Length comparisons examine the interaction of sequence length with delay (Tasks 4--5), or the effect of sequence length at fixed target recentness (Tasks 4--5). In all cases, intervals are adjusted for the nine comparisons in the crossed study. BA refers to balanced accuracy and FPR to false-positive rate.

\begin{table}[htbp]\centering\small
\caption{Planned comparisons in the crossed study. Differences are in percentage points; intervals are adjusted across all nine comparisons.}\label{supp:factorial_contrasts}
\input{tables/FACTORIAL_CONTRASTS.tex}\end{table}

\paragraph{Repeated use across trials.}
Each of nine experimental conditions uses forty trials. Photographs and objects do not repeat between trials within a bank, though banks may overlap. Within trials, objects may reappear with different renderings of the same photograph. Trials were balanced across positions using forty cyclic rotations; images were presented for one time unit, with a total of 64 blank units in Tasks 2/3. Comparisons across banks do not isolate the effects of delay or load.

\paragraph{Trial lengths and boundaries.}
We used sequences of eight images in Task 4 so that we could construct forty photograph-disjoint trials; Task 5 used eight- and sixteen-image sequences. Eight-unit boundaries advanced time and context, with normal memory decay and additional attenuation on the first image of each trial.

\paragraph{Carry versus reset.}
All three models used their fixed, source-calibrated response rules in both modes. In carry mode, $h$ and $F$ carry over from trial to trial, subject to decay and first-image attenuation; in reset mode, they are cleared. Both modes clear decision summaries at trial boundaries. Scores in carry mode change by between $-0.08$ and $+0.50$ percentage points between the first and last ten positions; scores in reset mode do not change (Table~\ref{supp:fixed40}).

The cyclic rotations are not statistically independent: they are different orderings of the same set of forty trials, and context is linked to trial position within a session. They are also not all possible orderings of the trials.

\begin{table}[htbp]\centering\small
\caption{Results with memory carried across trials or reset at the start of each trial (\%). Each of the forty trials appears at every position, and results average three models. Tasks 1/2/4/5 show balanced accuracy after pooling binary responses; Task 3 shows complete-trial success. First ten refers to trial positions 1--10 and last ten to 31--40. Reset scores are constant across positions. Changes are descriptive, computed as last minus first before rounding.}\label{supp:fixed40}
\input{tables/FIXED_CONDITION_40.tex}\end{table}

\section{Same-memory readout comparison}\label{supp:ordering}
\begin{figure}[htbp]\centering\includegraphics[width=\textwidth]{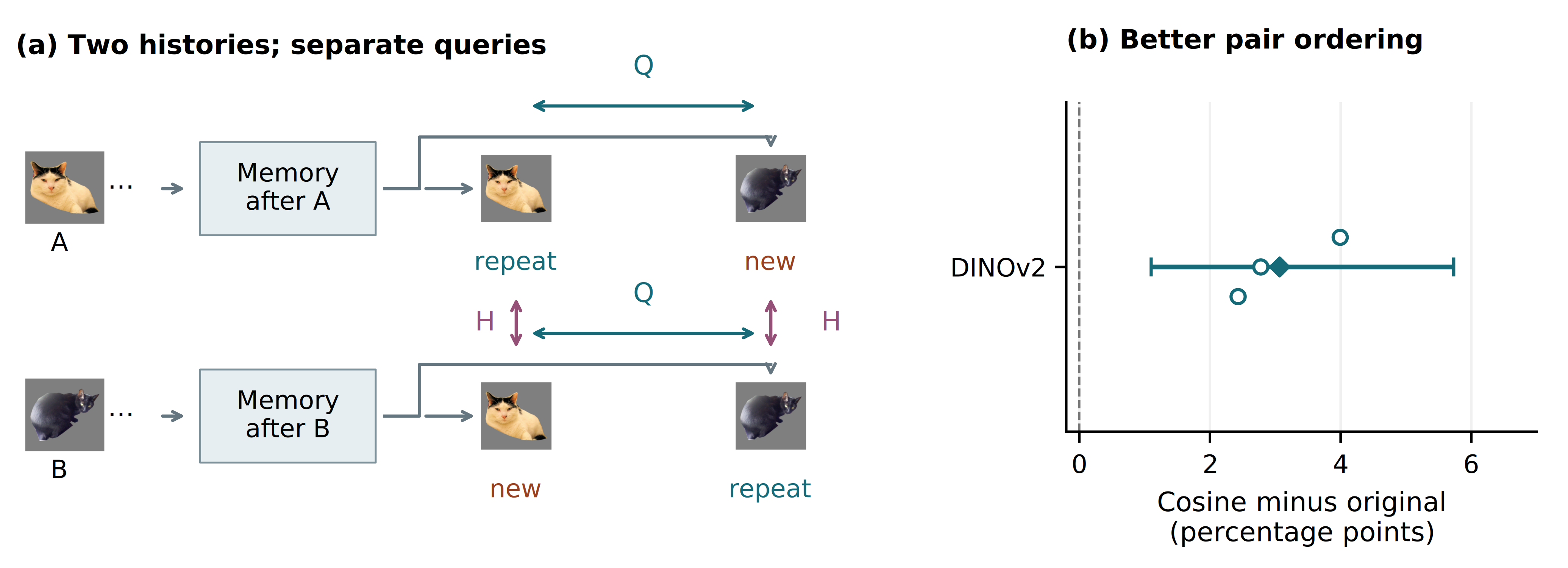}
\caption{Same-memory readout comparison. Each history was mirrored, changing only the earlier-presented object. The same query was repeated in one history and novel in the other. Within each history, each query and readout used an independent copy of the same aged memory state, before the query image was written. Horizontal comparisons ($Q$) assess paired ordering: does the repeated image score higher than the novel image? Vertical comparisons ($H$) assess history sensitivity: does the same image score higher after the history in which it repeated? Open circles show individual model means; the diamond shows their mean. The bar shows an exploratory 95\% confidence interval over 32 photograph groups. Images are illustrative.}\label{fig:protocol}
\end{figure}

To separate memory from readout~\citep{davari2022,anthes2023}, we created pairs of image histories that differed only in the identity of an earlier-presented object, A or B (Figure~\ref{fig:protocol}). The same query object was repeated after one history and novel after the other. Histories contained 7, 15 or 31 images before the query, with the earlier object appearing 1, 3 or 7 image positions before it; all other images, timings and instructions were identical.

For each history, queries and readouts used independent copies of memory saved before the query's write. We assessed paired ordering ($Q$: within each history, did the repeated image outscore the novel image?) and history sensitivity ($H$: did the same image score higher after the history in which it repeated?). Each measure averaged two comparisons, with half credit for ties; no ties occurred. There were 864 four-score cases across nine lag and history-length conditions, 32 photograph groups and three fixed models, with both mirrors included in each case.

Averaged over these cases, history sensitivity was high for the trained readout (99.71\%) and cosine similarity (99.83\%); paired ordering was 96.35\% and 99.42\%, respectively. Cosine similarity outperformed the trained readout by 3.07 percentage points on paired ordering [exploratory 95\% interval 1.10, 5.73], with every model showing an improvement. Of the 63 cases with imperfect paired ordering, 62 had perfect history sensitivity. Cosine similarity corrected 53 of the 63 ordering failures without introducing new failures. Neither metric directly reports recognition accuracy.

Confidence intervals resampled the 32 photograph groups used in previous analyses, with 10,000 iterations retaining all conditions, query copies and fixed models. They do not reflect uncertainty introduced by retraining. The five-task model was unchanged in these analyses: better ordering does not necessarily mean better YES/NO classification, and familiarity signals cannot substitute for matching a query to the designated sample.

\section{Code, data and image credits}\label{supp:reproduction}
\paragraph{Code and data availability.}
The code, trained model checkpoints, image feature banks and analysis materials are not publicly available with this version of the preprint. We plan to release these materials in a future version.

\paragraph{Image credits.}
Illustrative photographs are used under
\href{https://creativecommons.org/licenses/by/2.0/}{CC BY 2.0}:
\href{https://www.flickr.com/photos/dq090702/2794346421/}{\emph{White Cat}},
brokinhrt2;
\href{https://www.flickr.com/photos/50066720@N03/5713242946/}{\emph{Laptoplola}},
(: Rebecca-louise :);
\href{https://www.flickr.com/photos/thecraftyelephant/7743734120/}{\emph{Hat Red Velvet Front}},
Yarn Candy / Mistress\_Of\_Burden; and
\href{https://www.flickr.com/photos/vinothchandar/5911652454/}{\emph{The Reverse Blessing}},
VinothChandar. Images were cropped, masked and resized; the alternate cat image was also rotated. These are unscored illustrations.

%% file: tables/BATTERY_PROTOCOL.tex
\begin{tabular}{@{}llrrrl@{}}\toprule
Task & Panel & Trials & Binary & Positive/negative & Main schedule \\ \midrule
1 & Broad & 384 & 384 & 192/192 & delays 0,16,64,128,256,512 \\
2 & Broad & 608 & 608 & 304/304 & loads 0,1,4,8 \\
3 & Broad & 608 & 608 & 304/304 & loads 0,1,4,8; category + final \\
4 & Broad & 640 & 640 & 320/320 & lengths 4,8,16,32,64,128; terminal \\
5 & Broad & 248 & 8,960 & 3,890/5,070 & lengths 16,32,64,512 \\
\bottomrule\end{tabular}

%% file: tables/CONTROLLED_ENDPOINTS.tex
\begin{tabular}{@{}llrrrrrr@{}}\toprule
 & & \multicolumn{3}{c}{Intervening images} & \multicolumn{3}{c}{Equal-duration blanks} \\
Task & Length & Hit & FPR & BA & Hit & FPR & BA \\ \midrule
4 & 4 & 100.0 & 4.2 & 97.9 & 100.0 & 4.2 & 97.9 \\
4 & 8 & 100.0 & 13.9 & 93.1 & 100.0 & 4.2 & 97.9 \\
4 & 16 & 94.4 & 26.4 & 84.0 & 100.0 & 4.2 & 97.9 \\
4 & 32 & 86.1 & 59.7 & 63.2 & 100.0 & 5.6 & 97.2 \\
4 & 64 & 95.8 & 91.7 & 52.1 & 100.0 & 5.6 & 97.2 \\
\midrule
5 & 4 & 100.0 & 15.3 & 92.4 & 100.0 & 9.7 & 95.1 \\
5 & 8 & 100.0 & 11.1 & 94.4 & 100.0 & 9.7 & 95.1 \\
5 & 16 & 97.2 & 19.4 & 88.9 & 100.0 & 9.7 & 95.1 \\
5 & 32 & 86.1 & 31.9 & 77.1 & 100.0 & 9.7 & 95.1 \\
5 & 64 & 86.1 & 51.4 & 67.4 & 100.0 & 11.1 & 94.4 \\
\bottomrule\end{tabular}

%% file: tables/FACTORIAL_CONTRASTS.tex
\begin{tabular}{@{}lrr@{}}\toprule
Planned comparison & Difference (pp) & Adjusted interval \\ \midrule
Task 1: viewing time $\times$ delay & -4.17 & [-12.50, 0.00] \\
Task 2: count $\times$ delay, sample BA & +1.39 & [0.00, +4.17] \\
Task 2: count $\times$ delay, familiar FPR & -6.94 & [-18.06, 0.00] \\
Task 3: count $\times$ delay, sample BA & +1.39 & [0.00, +4.17] \\
Task 3: count $\times$ delay, all-correct & 0.00 & [0.00, 0.00] \\
Task 4: length $\times$ delay & +1.39 & [-2.08, +4.86] \\
Task 5: length $\times$ delay & -2.08 & [-10.42, +6.25] \\
Task 4: length at fixed recentness & -17.36 & [-31.25, -3.47] \\
Task 5: length at fixed recentness & -15.97 & [-23.84, -6.71] \\
\bottomrule\end{tabular}

%% file: tables/FIXED_CONDITION_40.tex
\begin{tabular}{@{}lrrrr@{}}\toprule
Fixed condition & Carry: first ten & Carry: last ten & Change (pp) & Reset \\ \midrule
Task 1: delay 16 & 94.25 & 94.17 & -0.08 & 93.33 \\
Task 1: delay 64 & 92.17 & 92.33 & +0.17 & 91.67 \\
Task 2: one distractor & 92.83 & 93.17 & +0.33 & 90.83 \\
Task 2: four distractors & 92.92 & 93.42 & +0.50 & 88.33 \\
Task 3: one distractor & 82.17 & 82.08 & -0.08 & 83.33 \\
Task 3: four distractors & 44.58 & 44.58 & +0.00 & 45.00 \\
Task 4: eight images & 90.00 & 90.17 & +0.17 & 89.17 \\
Task 5: eight images & 98.91 & 98.94 & +0.03 & 98.79 \\
Task 5: sixteen images & 96.91 & 96.91 & +0.01 & 96.93 \\
\bottomrule\end{tabular}